\documentclass[10pt,twocolumn,letterpaper]{article}
\usepackage[margin=0.72in]{geometry}
\usepackage{times}
\usepackage{graphicx}
\usepackage{amsmath,amssymb,amsfonts}
\usepackage{booktabs}
\usepackage[numbers,sort&compress]{natbib}
\usepackage{xcolor}
\usepackage[font=small,labelfont=bf]{caption}
\usepackage{subcaption}
\usepackage{url}
\usepackage{tabularx}
\usepackage{multirow}
\usepackage{array}
\usepackage{placeins}
\newcolumntype{Y}{>{\raggedright\arraybackslash}X}
\newcommand{\blank}{\mbox{[\kern0.4em]}}

\newlength{\tableedgepad}
\usepackage{pifont}
\newcommand{\cmark}{\ding{51}}
\newcommand{\xmark}{\ding{55}}

\definecolor{paperblue}{rgb}{0.21,0.49,0.74}
\usepackage[pagebackref,breaklinks,colorlinks,allcolors=paperblue]{hyperref}
\title{LongEarth-Seek: Keeping Long-Horizon Remote Sensing Reasoning on Track through Process-Guided Evidence Seeking}
\author{
Yupan Ding$^{1}$, Jing Xiao$^{1}$, Zhenyuan Zhang$^{2}$, Chaofeng Chen$^{1}$, Liang Liao$^{3}$, Mi Wang$^{4,5}$\\[0.4em]
{\small $^{1}$School of Artificial Intelligence, Wuhan University, Wuhan 430072, China}\\
{\small $^{2}$School of Computer Science, Wuhan University, Wuhan 430072, China}\\
{\small $^{3}$Hangzhou Research Institute, Xidian University, Hangzhou 311231, China}\\
{\small $^{4}$State Key Laboratory of Information Engineering in Surveying, Mapping and Remote Sensing, Wuhan University, Wuhan 430072, China}\\
{\small $^{5}$Hubei Luojia Laboratory, Wuhan 430079, China}
}
\begin{document}
\maketitle
\begin{abstract}
	Understanding long-term surface change from remote sensing imagery requires identifying critical observations and maintaining coherent reasoning across extended temporal sequences. However, existing approaches face two fundamental challenges: attention dispersion, where sparse but decisive change evidence is diluted by redundant or seasonal observations, and attention drift, where repeated inspections can shift the search away from the target region or the original task. To address these challenges, we propose LongEarth-Seek, a vision-language agent for process-guided active evidence seeking. Specifically, Dual-Path Temporal Selection identifies critical change observations through relation-guided temporal comparisons, distinguishing meaningful transitions from seasonal variations. Same-region Tracking anchors observations across dates to a consistent geographic target, preventing spatial drift during long-horizon evidence seeking. These components are integrated through Process-guided Memory, which maintains verified and unresolved change relations to guide subsequent observations and progressively refine the evolving interpretation. We further introduce LES-Bench, a benchmark covering diverse land-cover types and long-term change dynamics across four progressive reasoning levels. Extensive experiments demonstrate that LongEarth-Seek effectively identifies critical change stages and actively seeks missing observations to refine its interpretation.
	
\end{abstract}
\section{Introduction}
\label{sec:intro}
\begin{figure*}[!t]
	\centering\includegraphics[width=0.99\textwidth]{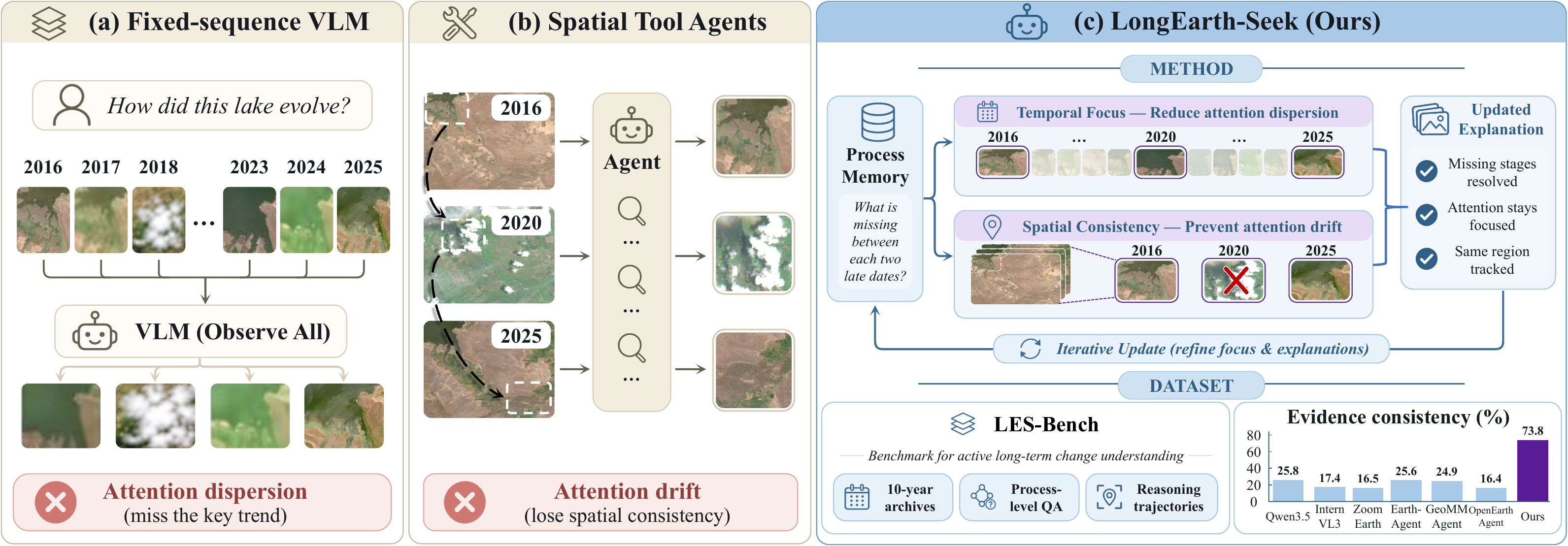}
	\vspace{-1mm}
	\caption{\textbf{From fixed-image understanding to active long-term process reasoning.} Fixed-sequence VLMs must interpret supplied images with redundant frames and unrelated changes. Spatial tool agents reveal local detail, with limited support for following change over long periods. LongEarth-Seek uses process memory to identify unresolved relations, select temporal comparisons, and track a target through local crops. These views connect key stages of its development.}
	\vspace{-3mm}
	\label{fig:motivation}
\end{figure*}

Satellite image time series record how cities, forests, and water bodies change over years~\cite{sentinel2,jrcwater,hansen}. Understanding these changes requires identifying not only what has changed, but also when changes occurred, how they unfolded, and whether they persisted or reversed. Similar changes observed at two dates may result from different histories, such as gradual development, abrupt disturbance, or recovery. Distinguishing these processes requires reconstructing intermediate stages and their temporal relationships. However, the decisive evidence may be sparsely distributed across years of imagery, making long-term change understanding more than simply comparing available observations.

Remote-sensing vision-language models (VLMs) have progressed from describing changes between image pairs~\cite{levircc,changeagent} to reasoning over images from multiple dates~\cite{teochat,earthdial,dynamicvl,geochrono}. However, long-horizon change understanding still faces two fundamental challenges. The first is \textbf{attention dispersion}: decisive evidence revealing key change stages may be diluted by redundant, cloud-obscured, or seasonally varying views, as models distribute attention across long image sequences rather than focusing on critical changes. Recent remote-sensing agents incorporate region inspection and tool use~\cite{zoomearth,earthagent}, enabling models to acquire additional observations beyond fixed inputs. However, prolonged observation introduces a second challenge, \textbf{attention drift}: successive reasoning and observation decisions may gradually deviate from the original change question or geographic target, leading to inconsistent evidence across dates. Consequently, models may inspect many images yet fail to identify or connect the observations needed to explain long-term changes.

Addressing these challenges requires actively seeking decisive evidence while maintaining reasoning consistency across space and time. Importantly, the value of a view depends not merely on the magnitude of visual change, but on whether it resolves uncertainty about the ongoing change process. For example, identifying when a transition begins may require comparing neighboring dates, whereas verifying whether a change persists may require comparing views from similar seasons across years. Likewise, views from different dates must remain anchored to the same geographic target to support a coherent interpretation. We therefore hypothesize that unresolved change relations can guide the search for critical evidence, while new views can refine the process interpretation and inform subsequent searches. This creates a closed loop between active evidence seeking and long-term change understanding.

We propose \textbf{LongEarth-Seek}, a vision-language agent for process-guided active evidence seeking in long-term surface change understanding (Fig.~\ref{fig:motivation}(c)). Rather than treating all views equally or making independent tool requests, LongEarth-Seek maintains an evolving process state to guide observation decisions and interpretation updates. Specifically, \textit{Dual-Path Temporal Selection} identifies decisive temporal evidence based on unresolved change relations, combining adjacent-date comparisons to localize transitions with season-aligned comparisons to distinguish persistent changes from seasonal variations. \textit{Same-region Tracking} maintains geographic consistency across dates, ensuring that views remain anchored to the same target throughout long-horizon searches. \textit{Process-guided Memory} integrates these capabilities by tracking verified and unresolved change relations, enabling the agent to seek missing evidence, revise its interpretation, and choose what to inspect next. Furthermore, \textit{Joint State Supervision} jointly trains observation decisions, process-state updates, and final answers to encourage coherent, evidence-grounded reasoning throughout the search.


To evaluate long-term change understanding and active spatiotemporal evidence search, we introduce \textbf{LES-Bench}, comprising 10,340 questions across 600 geographic regions worldwide. Unlike conventional benchmarks that focus on reasoning over fixed image sequences, LES-Bench evaluates how models identify and acquire missing spatial and temporal evidence to explain long-term surface changes. It covers four task levels, ranging from interpreting provided views to actively seeking missing evidence and reconstructing intermediate stages of evolution. By varying the availability of observations and the requirements for process reasoning, LES-Bench evaluates whether models can find key evidence and maintain consistent observations across long-term sequences. It therefore provides a unified evaluation of evidence acquisition and long-term change understanding. Extensive experiments demonstrate that LongEarth-Seek effectively identifies critical change stages and actively seeks missing evidence to refine its interpretation.

Our contributions are summarized as follows.

\begin{itemize}
	\item  \textbf{Paradigm}. We introduce a process-guided active evidence-seeking paradigm for long-term surface change understanding, addressing attention dispersion and drift through targeted evidence acquisition and coherent spatiotemporal reasoning.
	
	\item \textbf{Methodology}. We propose LongEarth-Seek, a vision-language agent that jointly learns evidence acquisition and process reasoning to identify critical change stages, maintain spatiotemporal consistency, and progressively refine change interpretations.
	
	\item \textbf{Benchmark and validation}. We introduce LES-Bench, covering diverse land-cover types and long-term change dynamics across four progressive reasoning levels. Extensive experiments demonstrate and validate the agent's ability to identify critical change evidence, sustain evidence-grounded reasoning, and generalize to broader geospatial tasks.
	
	
	
\end{itemize}

\section{Related Work}
\label{sec:related}
\noindent\textbf{Long-term remote-sensing understanding.}
Remote-sensing models learn multimodal Earth representations~\cite{skysense,terramind,olmoearth,ramen,robsense}, align images with text~\cite{remoteclip,geochat,skysenseo}, and link language to image regions~\cite{terrascope,segearthr2}. Prior work models satellite time series~\cite{vitssits}, detects change~\cite{spectralchange}, and describes it in language~\cite{levircc,changeagent}. TEOChat and EarthDial extend instruction following to temporal imagery~\cite{teochat,earthdial}. DynamicVL and GeoChrono use multi-year imagery to study development and long-term reasoning~\cite{dynamicvl,geochrono}.

\noindent\textbf{Active visual agents.}
Visual agents compose tools to gather evidence~\cite{visprog,vipergpt}. In remote sensing, they use image, GIS, and spectral tools~\cite{rsagent,earthagent,openearthagent,geommag,hirsagent}, actively inspecting high-resolution regions~\cite{zoomearth}. Long-video methods select clips or retrieve evidence~\cite{act2see,activevideoperception,longvideor1,drvideo,salova,videotree} and organize it across time~\cite{timechat,moviechat,rewind,videoarm,svagent}. LongEarth-Seek extends these ideas to long satellite sequences. It uses unresolved relations in a target's history to choose the next temporal comparison and revise the process state.

\noindent\textbf{Remote-sensing benchmarks.}
TEOChatlas evaluates temporal instruction following; DVL-Bench and ChronoBench examine multi-year development and temporal reasoning over supplied imagery~\cite{teochat,dynamicvl,geochrono}. LRS-GRO tests active regional perception~\cite{zoomearth}. GeoMMBench, Earth-Bench, and UnivEARTH cover broader geospatial or tool-assisted tasks~\cite{geommag,earthagent,univearth}. Video benchmarks assess reasoning over long sequences~\cite{videomme,longvideobench,lvbench,mlvu,longvale}. Table~\ref{tab:benchmarks} shows how LES-Bench tests tool use to select dates, track the same parcel, and recover missing stages.
\begin{table}[!t]
	\centering\setlength{\tabcolsep}{13pt}
	\caption{Capabilities of LES-Bench and related remote-sensing benchmarks. LT: images from at least three distinct dates; ST: reasoning about intermediate stages and their temporal relations; RG: identifying the image region corresponding to the target; TS: actively selecting additional dates to find evidence; T: using tools to inspect images or obtain analysis results. \cmark{} indicates coverage and \xmark{} indicates no coverage in the reported evaluation.}
	\label{tab:benchmarks}
	\resizebox{\columnwidth}{!}{%
		\begin{tabular}{@{\hspace{5pt}}lccccc@{\hspace{5pt}}}
			\toprule
			Benchmark & LT & ST & RG & TS & T\\
			\midrule
			TEOChatlas~\cite{teochat} & \cmark{} & \cmark{} & \cmark{} & \xmark{} & \xmark{}\\
			DVL-Bench~\cite{dynamicvl} & \cmark{} & \cmark{} & \cmark{} & \xmark{} & \xmark{}\\
			ChronoBench~\cite{geochrono} & \cmark{} & \cmark{} & \cmark{} & \xmark{} & \xmark{}\\
			GeoMMBench~\cite{geommag} & \cmark{} & \xmark{} & \cmark{} & \xmark{} & \cmark{}\\
			LRS-GRO~\cite{zoomearth} & \xmark{} & \xmark{} & \cmark{} & \xmark{} & \cmark{}\\
			Earth-Bench~\cite{earthagent} & \cmark{} & \xmark{} & \cmark{} & \xmark{} & \cmark{}\\
			UnivEARTH~\cite{univearth} & \cmark{} & \xmark{} & \xmark{} & \xmark{} & \cmark{}\\
			OpenEarthAgent~\cite{openearthagent} & \xmark{} & \xmark{} & \cmark{} & \xmark{} & \cmark{}\\
			\midrule
			LES-Bench & \cmark{} & \cmark{} & \cmark{} & \cmark{} & \cmark{}\\
			\bottomrule
	\end{tabular}}
	\vspace{-2mm}
\end{table}

\FloatBarrier

\section{LES-Bench}
\label{sec:benchmark}
\subsection{Change Process Collection}
\label{sec:curation}

\begin{figure*}[t]
	\centering\includegraphics[width=0.99\textwidth]{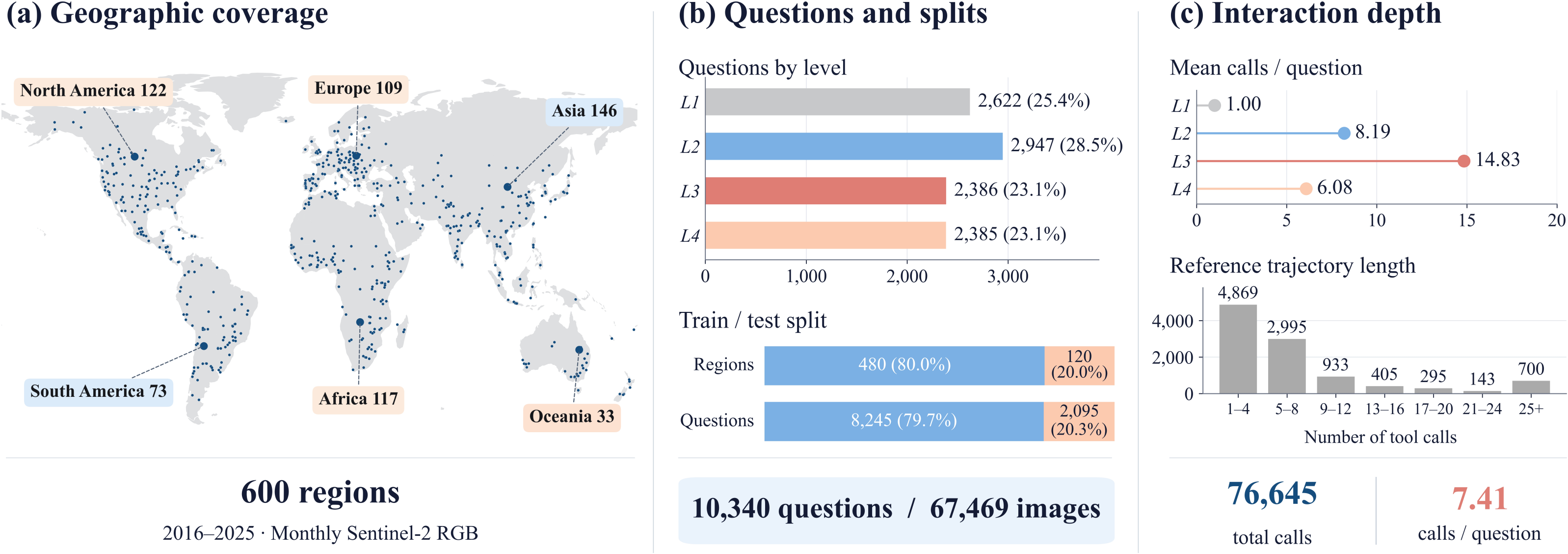}
	\vspace{-2mm}
	\caption{\textbf{LES-Bench composition and interaction demands.} (a) Geographic distribution of the 600 regions. (b) Question counts across \textit{L1}--\textit{L4} and the region-level train/test split. (c) Mean tool calls per question at each level and the distribution of reference trajectory lengths.}
	\vspace{-4mm}
	\label{fig:dataset}
\end{figure*}

Long-term satellite sequences contain seasonal variations, unrelated changes, and cloud- or shadow-affected observations, making it necessary to distinguish meaningful processes from incidental visual differences. We collect 67,469 monthly Sentinel-2 RGB images from 600 regions during 2016--2025~\cite{sentinel2}, each covering $6\times6$\,km at 10\,m resolution. Available months retain an acquired scene, while missing months remain absent. Figure~\ref{fig:dataset} summarizes the dataset.

Aligned land-cover estimates and spectral measurements identify candidate changes, supported by clear-pixel masks and additional water and forest products~\cite{dynamicworld,jrcwater,hansen}. We inspect the dated images to verify the target, change direction, and intermediate states. Verified changes are organized into process records linking these states by time and spatial extent. Thus, auxiliary products guide candidate discovery, while observations determine the process labels.

\subsection{Questions and Process Annotations}
\label{sec:process_qa}
\noindent\textbf{Question construction.}
A process record describes where a target changes, what changes, and how its visible states are related over time. Its structure depends on the process. Construction may involve site preparation, active building, and completed development. Vegetation can undergo disturbance and recovery; water extent can expand, retreat, or recur; agricultural land can follow a seasonal cycle. The reference preserves the stages needed to answer a particular question instead of imposing one start--progress--completion pattern on every process.

Questions turn these records into tasks by specifying the requested finding and the information available at the start. A question about the onset of construction requires dates that bound the transition. A question about gradual development requires intermediate stages as well. The task wording preserves the recorded target and process, while the initial evidence determines what the model must search for. The answer states the supported target, change, interval, and stages in natural language. It reports an interval when the images cannot establish an exact date.

\noindent\textbf{Process annotations.}
The process ground truth provides the basis for checking both the answer and the search. It associates each required stage with a location, a date or valid date range, and supporting image identifiers. Temporal relations record which stages precede others or whether a state persists or recurs. Equivalent dates and alternative locations are retained when their images support the requested process. Thus a valid search can differ from a single reference path. For example, a construction record links images of open land in 2018, site preparation in 2021, and buildings in 2023 at the same parcel. These stages provide the reference for scoring reconstruction.

Process relations require more care than a final change category: an image can show a building without establishing when construction began. Of 11,656 source records, 10,340 meet the benchmark's evidence sufficiency criteria. Two annotators independently reviewed a stratified sample of 1,166 records, checking the target, direction, dated stages, and relations against the images. Raw agreement was 91.2\%, with mean Cohen's $\kappa=0.85$. A senior reviewer resolved disagreements by revisiting the images and supporting sources. The remaining 1,316 source records are outside the train/test split.

\subsection{Four-Level Task Hierarchy}
\label{sec:task}
The same construction record illustrates how the initial evidence changes the task. (1) \textit{L1} (supplied-image interpretation) gives the parcel and the relevant images and asks what changed; (2) \textit{L2} (time or location search) gives either the parcel and a time range and asks when construction began (\textit{L2-T}), or dated images and asks where construction appeared (\textit{L2-S}); (3) \textit{L3} (joint discovery) gives a wider region and a time range and asks where and when construction began; (4) \textit{L4} (process reconstruction) gives the parcel and time range, sometimes with endpoint images. The model traces how the site developed by finding intermediate stages, such as site preparation, and linking them to the initial and final states.

\begin{figure*}[t]
	\centering\includegraphics[width=0.99\textwidth]{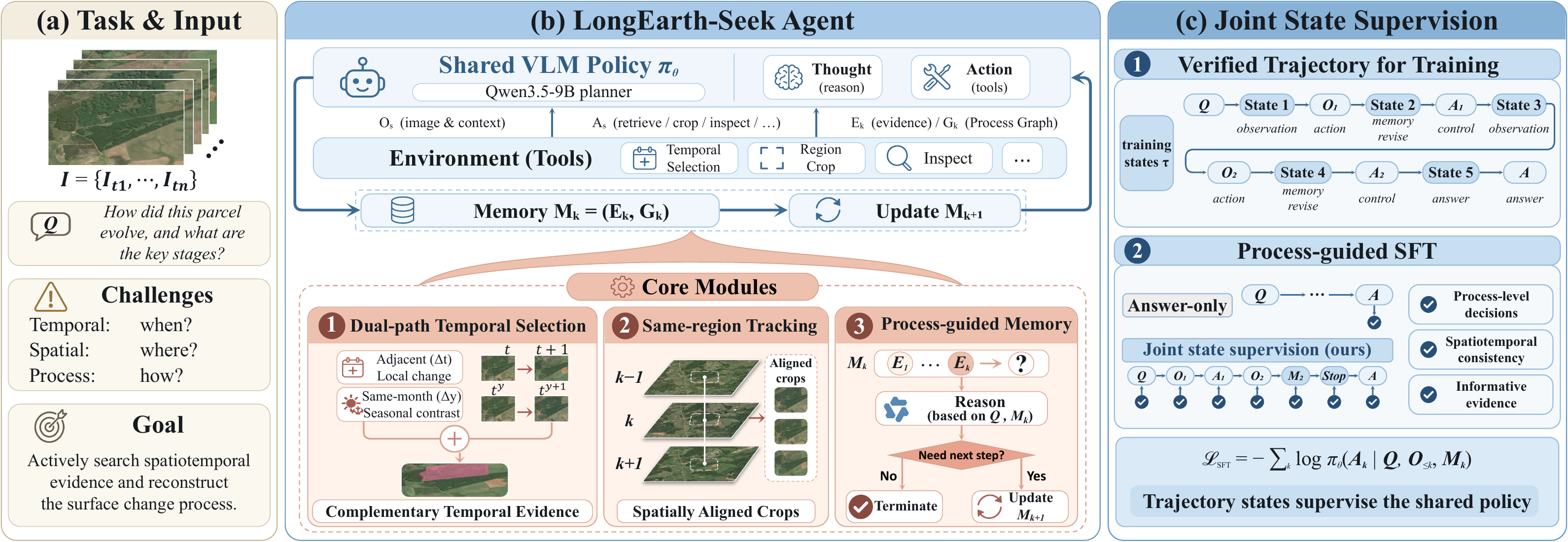}
	\vspace{-1mm}
	\caption{\textbf{LongEarth-Seek framework.} (a) A question and regional image sequence define the search task. (b) A shared Qwen3.5-9B policy requests dated views through tools and updates its process state from the returned views. Dual-Path Temporal Selection provides adjacent-date and same-month comparisons. Same-region Tracking aligns views across dates. Process-guided Memory guides further inspection or stopping. (c) Joint State Supervision trains the policy to select views, update the process state, and answer.}
	\label{fig:framework}
	\vspace{-2mm}
\end{figure*}

LES-Bench contains 2,622 \textit{L1}, 2,947 \textit{L2}, 2,386 \textit{L3}, and 2,385 \textit{L4} questions. Reference evidence uses one date for 495 questions, two for 1,921, and at least three for 7,924; 5,859 use six or more. These counts concern reference evidence, not initial model inputs. Geographic regions and shared source events remain within one split, with a 25\,km train/test separation. The split contains 480 training regions with 8,245 questions and 120 test regions with 2,095 questions. The primary benchmark has sufficient image evidence for its answers; a separate masked test evaluates recognition of missing evidence.

\subsection{Interaction Trajectory Construction}
\label{sec:trajectories}
Question--answer pairs alone do not teach an agent when to inspect another date, change a candidate location, or revise an explanation. We construct interaction trajectories from the training process references. Using the known training targets, the construction procedure selects dates and local regions whose images can test the required stages and relations. It executes these requests through the observation tools and checks the returned evidence.

The selected actions are then replayed from the task's initial state. The policy sees the public question and initial evidence, followed only by images returned up to the current step. Each observation is paired with a process-state update and a decision to continue, revise the candidate, or answer. An uninformative view leaves the relevant question open; conflicting evidence can lead to a revised interpretation. Private references determine the supervision targets, but are absent from the policy's inference inputs.

Each complete trajectory yields multiple supervised states: choosing an observation, updating the process interpretation, controlling the search, and answering from the collected evidence. The 8,245 training questions yield 38,246 such states. They train one shared Qwen3.5-9B policy for both tool use and process reasoning, linking data construction to Joint State Supervision (Sec.~\ref{sec:learning}).

\section{LongEarth-Seek}
\label{sec:method}

LongEarth-Seek employs a shared VLM policy to reason about change and select observations (Fig.~\ref{fig:framework}). Process-guided Memory tracks the current interpretation and unresolved evidence gaps, guiding Dual-Path Temporal Selection to identify complementary time comparisons and Same-region Tracking to maintain spatial consistency across dates. Given the question, initial images, available dates, and tool descriptions, the policy selects temporal and spatial observations to resolve missing evidence. Each returned view is used to update the process state, informing subsequent searches or the final answer. When essential stages remain unverifiable, the policy can express uncertainty. This loop adapts observation decisions to available evidence and the evolving interpretation.

\subsection{Dual-Path Temporal Selection}
\label{sec:temporal_selection}
For a question $q$, let $\mathcal T_g=(t_1,\ldots,t_n)$ be the available dates for region $g$, ordered in time. We form adjacent-date comparisons and same-month comparisons across years:
\begin{equation}
	\begin{aligned}
		\mathcal P_{\mathrm{adj}} &= \{(t_i,t_{i+1})\}_{i=1}^{n-1},\\
		\mathcal P_{\mathrm{yr}} &= \{(t_i,t_j): i<j,\;m(t_i)=m(t_j),\\
		&\hspace{20mm} y(t_i)\ne y(t_j)\}.
	\end{aligned}
	\label{eq:temporal_paths}
\end{equation}
The two paths answer different questions about change. Adjacent dates help order nearby stages and narrow when a change began. Same-month comparisons across years help distinguish lasting change from seasonal variation. An unresolved onset or stage order often calls for adjacent dates, while persistence or seasonal recurrence often calls for a same-month comparison. The VLM weighs both paths against the question and current process state, so its choice follows the relation to be resolved.

\subsection{Same-region Tracking}
\label{sec:region_tracking}
For an open-search question, the VLM first proposes a candidate parcel from the available regional views. It checks the candidate against the question and the returned local images; a candidate that does not show the requested process is set aside, and the search continues. Once a parcel $b$ is selected, the agent uses a shared geographic window $B_b$ at every chosen date:
\begin{equation}
	C_t^b = \operatorname{Crop}(I_{g,t},B_b), \qquad t\in\mathcal T_g.
	\label{eq:shared_crop}
\end{equation}
The shared crop provides geographically anchored temporal tracking at parcel scale. It keeps local views tied to the same place while preserving enough surroundings to reveal boundary change. If the target grows beyond the current window, the policy can expand it around the geographic anchor and use the expanded bounds for later comparisons. 

\begin{table*}[t]
	\centering\small\setlength{\tabcolsep}{3pt}
	\caption{\textbf{Model comparison on the complete 2,095-question LES-Bench test set.} \textit{L1}--\textit{L4} and Overall report AQ, the percentage of fully correct answers. ESPA requires sufficient visibility in the actual model input, support from inspected evidence, and independent source validation. All scores are percentages. $\uparrow$ indicates that higher is better. Bold and underline mark the best and second-best results.}
	\vspace{-2mm}
	\label{tab:main}
	\begin{tabular}{@{\hspace{\tableedgepad}}>{\raggedright\arraybackslash}p{1.34in}>{\raggedright\arraybackslash}p{1.64in}*{4}{>{\centering\arraybackslash}p{0.41in}}>{\centering\arraybackslash}p{0.64in}>{\centering\arraybackslash}p{0.67in}@{\hspace{\tableedgepad}}}
		\toprule
		Model / method & Backbone / controller & \textit{L1}\,$\uparrow$ & \textit{L2}\,$\uparrow$ & \textit{L3}\,$\uparrow$ & \textit{L4}\,$\uparrow$ & Overall\,$\uparrow$ & ESPA\,$\uparrow$\\
		\midrule
		Qwen3.5-9B (Base)~\cite{qwen35} & Qwen3.5-9B & \underline{60.8} & 58.0 & \underline{35.7} & \underline{33.4} & \underline{47.92} & \underline{25.8} \\
		InternVL3-8B~\cite{internvl3} & InternVL3-8B & 49.6 & 36.8 & 25.0 & 18.6 & 33.22 & 17.4 \\
		\midrule
		ZoomEarth-3B~\cite{zoomearth} & ZoomEarth-3B & 38.1 & 24.1 & 2.3 & 19.4 & 21.58 & 16.5 \\
		Earth-Agent~\cite{earthagent} & Qwen3.5-9B & 60.1 & \underline{59.4} & 30.9 & 32.8 & 46.87 & 25.6 \\
		GeoMMAgent~\cite{geommag} & Qwen3.5-9B & 54.2 & 56.2 & 28.3 & 32.4 & 43.72 & 24.9 \\
		OpenEarthAgent~\cite{openearthagent} & OpenEarthAgent-4B & 52.2 & 43.2 & 21.5 & 25.5 & 36.42 & 16.4 \\
		\textbf{LongEarth-Seek (Ours)} & Qwen3.5-9B & \textbf{86.4} & \textbf{84.1} & \textbf{76.0} & \textbf{61.2} & \textbf{77.6} & \textbf{73.8} \\
		\bottomrule
	\end{tabular}
	\vspace{-2mm}
\end{table*}

\subsection{Process-guided Memory}
\label{sec:process_memory}
At inference, the model does not receive the complete reference process. It builds a provisional process graph from the question and the views it has inspected. Let $E_k$ denote the georeferenced views inspected by step $k$. We define the process state as
\begin{equation}
	\begin{aligned}
		M_k &= (E_k,\widehat{\mathcal G}_k),\\
		\widehat{\mathcal G}_k &= (\widehat{\mathcal S}_k,\widehat{\mathcal R}_k,\widehat{z}_k).
	\end{aligned}
	\label{eq:online_state}
\end{equation}
where $\widehat{\mathcal S}_k$ contains proposed surface stages, $\widehat{\mathcal R}_k$ contains the relations between them, and $\widehat{z}_k$ assigns each relation a status: open, supported, conflicted, or unverifiable. The model predicts and revises this state using only $q$ and $E_k$. The reference graph $\mathcal G_q^\star$ is used to supervise training and score predictions, but is not available to the policy at test time.

The current state makes the next decision specific to the process under study. Let $\widehat U_k$ denote the open or conflicted relations. The VLM selects one of them, then chooses a date comparison and geographic view that can test it:
\begin{equation}
	\begin{aligned}
		r_k &\sim \pi_\theta^R(\cdot\mid q,M_k),\qquad r_k\in\widehat U_k\cup\{\mathrm{stop}\},\\
		a_k &\sim \pi_\theta^A(\cdot\mid q,M_k,r_k),\qquad r_k\ne\mathrm{stop},\\
		M_{k+1} &= U_\theta\!\left(q,M_k,\operatorname{Tool}(a_k)\right).
	\end{aligned}
	\label{eq:relation_policy}
\end{equation}
The policy answers when $r_k=\mathrm{stop}$. If a required stage remains unseen but a useful date is available, the policy continues searching. New evidence can support a relation or revise it when it conflicts with the current explanation. If no available view can resolve an essential stage, the policy can report uncertainty. Consider a parcel shown as open land in 2018 and built in 2024. The endpoints do not reveal whether work began early and progressed gradually or started later and advanced quickly. The memory marks construction onset as unresolved, prompting the VLM to inspect an intermediate date that distinguishes these accounts. The returned view narrows the onset and updates the stage relations. In this way, process memory links the evidence already gathered to the evidence still needed by making the unresolved relation the target of the next observation.

\subsection{Joint State Supervision}
\label{sec:learning}
We adapt a shared Qwen3.5-9B policy~\cite{qwen35} with language-side QLoRA~\cite{qlora,lora}. Joint State Supervision links view selection, process state updates, and the final answer in one trajectory. The current state also guides whether to continue, change the candidate, or stop. The offline reference supplies supervision targets, while the policy sees only the public task, current process state, and views collected so far.

We balance supervision across the four task levels with a state-level SFT objective:
\begin{equation}
	\mathcal L_{\mathrm{SFT}}=\frac{1}{4}\sum_{\ell=1}^{4}\mathbb{E}_{q\sim\mathcal Q_\ell}\left[\ell_\theta(\tau_q)\right],
	\label{eq:balanced_sft}
\end{equation}
where $\mathcal Q_\ell$ is the set of questions at level $\ell$, and $\tau_q$ contains observation choices, process-state updates, control decisions, and the answer. The loss $\ell_\theta$ combines the supervised targets within a question, with the four levels receiving equal total weight.

\section{Experiments}
\label{sec:experiments}
\subsection{Experimental Setup}
\noindent\textbf{Evaluation protocol.}
We evaluate on 2,095 LES-Bench questions from 120 held-out regions. We train Qwen3.5-9B with QLoRA on 4 RTX 5090 GPUs (Sec.~\ref{sec:learning}). We use the same agent configurations on LES-Bench, VLRS-Bench, and GeoMMBench. ZoomEarth uses its released vision model. Earth-Agent and GeoMMAgent use Qwen3.5-9B for control and visual reasoning while retaining their own decision flows. OpenEarthAgent keeps its released 4B text controller and uses Qwen3.5-9B only to interpret images. On LES-Bench, all systems use the same image pool, tool interface, and search budget. For zero-shot transfer, all methods receive the same questions and images within each benchmark and follow its official scoring protocol. We use the released $512\times512$ images in their original order on VLRS-Bench~\cite{vlrsbench} and score multiple-choice answers on GeoMMBench~\cite{geommag}. Neither benchmark is used for fine-tuning or checkpoint selection.

\noindent\textbf{Evaluation metrics.} Answer quality (AQ) is the share of questions for which all required facts are confirmed. An untuned Qwen3.5-9B transcribes anonymized responses into structured claims. Quotes from the original response verify the extraction. A fixed set of rules then checks the claims against remote-sensing sources. \textit{L1} checks interpretation of supplied images. \textit{L2} checks the requested time or location. \textit{L3} checks where and when a change occurred and the relations required by the question. \textit{L4} checks the main stages and their order. The scorer accepts any dates, regions, and equivalent wording that satisfy the question. Each question receives one point if all required facts are confirmed and zero otherwise. Incorrect, incomplete, non-factual, and unresolved answers all remain in the denominator.

ESPA measures whether an answer can be verified from the evidence. For question $i$, the claimed dates and regions must be sufficiently visible in the actual input ($V_i$), the cited views must support the claims ($S_i$), and an independent source check must confirm them within the time range required by the question ($H_i$). A question counts only if all three checks pass. With $N$ test questions, we compute
\begin{equation}
	\mathrm{ESPA} = \frac{1}{N}\sum_{i=1}^{N}\mathbf{1}[V_i\land S_i\land H_i].
	\label{eq:espa_auto}
\end{equation}
AQ and ESPA are computed independently and reported as percentages. The supplement will report agreement with independent human ratings on a subset and identify verification sources independent of the reference annotations.

\subsection{Benchmark Results}
\label{sec:main_results}
\noindent\textbf{Results on LES-Bench.}
All models in Table~\ref{tab:main} have access to the same tools. The AQ gap is largest on \textit{L3}. Qwen3.5-9B (Base) scores 60.8\% on \textit{L1}, where evidence is supplied, but 35.7\% on \textit{L3} and 33.4\% on \textit{L4}. The other agents also lose accuracy when they must locate changes or reconstruct intermediate stages. LongEarth-Seek reaches 77.6\% overall AQ. Its ESPA of 73.8\%, compared with 25.8\% for the strongest baseline on this metric, shows how often its answers are backed by the views it inspected. LongEarth-Seek scores 76.0\% versus 35.7\% for the best baseline on \textit{L3}, and 61.2\% versus 33.4\% on \textit{L4}. Reconstructing intermediate stages remains hard after locating the change.

\noindent\textbf{Zero-shot transfer.}
On VLRS-Bench, LongEarth-Seek improves on Qwen3.5-9B in all four categories and raises the overall score from 32.2\% to 45.9\% (Table~\ref{tab:vlrs}). Its overall score is close to OpenEarthAgent (45.7\%) and above Earth-Agent (41.8\%) and GeoMMAgent (36.0\%). It also scores above GPT-4o and Gemini-3.1-Pro in all four categories. On GeoMMBench, it achieves 69.62\% accuracy, compared with 67.81\% for its base and 64.47\% for GeoMMAgent. The gains over the base extend to broader geospatial question answering.

\begin{table}[t]
	\centering\footnotesize\setlength{\tabcolsep}{0pt}
	\caption{\textbf{Zero-shot transfer on VLRS-Bench and GeoMMBench.} Both benchmarks use their official scoring protocols. Scores are percentages, with higher values indicating better performance. Cogn., Dec., and Pred. denote cognition, decision, and prediction; ST-C denotes spatiotemporal cognition. Overall is the benchmark's question-weighted score. GPT-4o and Gemini-3.1-Pro overall scores are estimated from rounded category scores. Dashes indicate unreported GeoMMBench results.}
	\vspace{-2mm}
	\label{tab:vlrs}
	\begin{tabular*}{\linewidth}{@{\extracolsep{\fill}}lcccccc@{}}
		\toprule
		& \multicolumn{5}{c}{VLRS-Bench} & {\scriptsize GeoMMBench}\\
		\cmidrule(lr){2-6}\cmidrule(l){7-7}
		Method & Cogn. & Dec. & Pred. & ST-C & Overall & Acc.\\
		\midrule
		Qwen3.5-9B~\cite{qwen35} & 32.1 & 32.3 & 29.9 & 34.5 & 32.2 & \underline{67.81}\\
		InternVL3-8B~\cite{internvl3} & 35.4 & 35.6 & 33.8 & 30.4 & 33.8 & 56.20\\
		EarthDial-4B~\cite{earthdial} & 28.6 & 26.9 & 28.0 & 26.2 & 27.4 & 42.62\\
		TEOChat~\cite{teochat} & 30.7 & 31.4 & 29.0 & 32.4 & 30.9 & 31.10\\
		GeoChrono~\cite{geochrono} & 42.1 & 41.8 & 41.0 & 42.2 & 41.8 & 59.74\\
		GPT-4o~\cite{gpt4o} & 38.0 & 31.0 & 34.8 & 36.9 & 35.5 & --\\
		Gemini-3.1-Pro~\cite{gemini31} & \underline{44.8} & 44.3 & 40.5 & 45.7 & 43.6 & --\\
		\midrule
		ZoomEarth-3B~\cite{zoomearth} & 29.2 & 26.7 & 27.6 & 27.6 & 27.8 & 46.75\\
		Earth-Agent~\cite{earthagent} & 44.2 & 45.7 & 34.5 & 42.9 & 41.8 & 35.33\\
		GeoMMAgent~\cite{geommag} & 35.9 & 35.2 & 32.3 & 40.4 & 36.0 & 64.47\\
		OpenEarthAgent~\cite{openearthagent} & 44.4 & \underline{46.2} & \textbf{45.0} & \underline{47.1} & \underline{45.7} & 47.64\\
		\textbf{LongEarth-Seek} & \textbf{44.9} & \textbf{46.4} & \underline{44.6} & \textbf{47.8} & \textbf{45.9} & \textbf{69.62}\\
		\bottomrule
	\end{tabular*}
	\vspace{-4mm}
\end{table}

\subsection{Ablations and Error Analysis}
\label{sec:mechanisms}
\label{sec:evidence_analysis}
\paragraph{Component ablation.}
Table~\ref{tab:ablation_main} tests how time selection, spatial tracking, and memory work together to keep the search focused. Memory gives the largest gain among single components. With memory in place, temporal selection gives the strongest pair on \textit{L3} (74.1\% AQ), while same-region tracking gives the strongest pair on \textit{L4} (57.5\% AQ) and ESPA (69.0\%). The full method improves on both pairs, consistent with the need to find informative dates and keep the views tied to the same target.

\begin{table}[t]
	\centering\setlength{\tabcolsep}{8pt}
	\caption{\textbf{Component ablation.} A: Dual-Path Temporal Selection; B: Same-region Tracking; C: Process-guided Memory. Missing A, B, or C uses generic selection, independent crops, or stage-only memory, respectively. \textit{L2}--\textit{L4} report AQ. Scores are percentages.}
	\vspace{-2mm}
	\label{tab:ablation_main}
	\resizebox{\columnwidth}{!}{%
		\begin{tabular}{lccccccc}
			\toprule
			Setting & A & B & C & \textit{L2} & \textit{L3} & \textit{L4} & ESPA\\
			\midrule
			Reference & \xmark{} & \xmark{} & \xmark{} & 61.5 & 42.0 & 36.8 & 30.6 \\
			\midrule
			A only & \cmark{} & \xmark{} & \xmark{} & 64.3 & 48.0 & 39.6 & 34.5 \\
			B only & \xmark{} & \cmark{} & \xmark{} & 65.0 & 46.5 & 42.9 & 37.2 \\
			C only & \xmark{} & \xmark{} & \cmark{} & 76.6 & 65.4 & 47.6 & 64.1 \\
			\midrule
			A + B & \cmark{} & \cmark{} & \xmark{} & 67.2 & 55.0 & 46.2 & 45.0 \\
			A + C & \cmark{} & \xmark{} & \cmark{} & \underline{81.2} & \underline{74.1} & 53.5 & 66.7 \\
			B + C & \xmark{} & \cmark{} & \cmark{} & 80.6 & 71.9 & \underline{57.5} & \underline{69.0} \\
			\midrule
			\textbf{Full method} & \cmark{} & \cmark{} & \cmark{} & \textbf{84.1} & \textbf{76.0} & \textbf{61.2} & \textbf{73.8} \\
			\bottomrule
	\end{tabular}}
	\vspace{-2mm}
\end{table}

\begin{figure}[t]
	\centering
	\includegraphics[width=0.99\columnwidth]{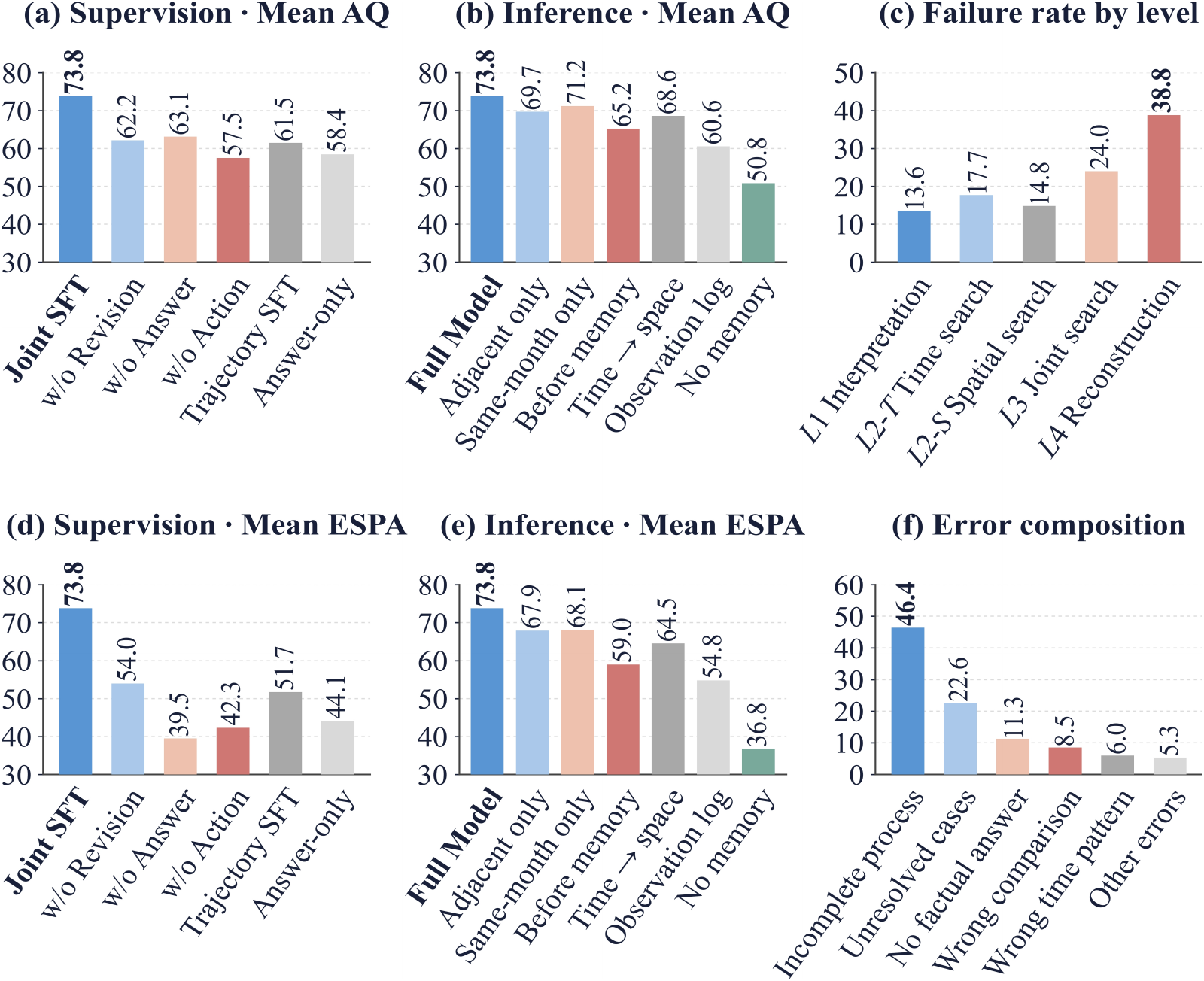}
	\vspace{-2mm}
	\caption{\textbf{Ablations and remaining failures in process-guided search.} Panels (a, b) show mean AQ and (d, e) show ESPA on \textit{L2}--\textit{L4}. \textit{L1} provides the required views and needs no search. Supervision ablations retrain the policy, while inference ablations change only the search strategy. On the full test set, (c) shows per-task failure rates ($1-\mathrm{AQ}$ with AQ as a fraction), and (f) groups the 470 failed answers by scoring outcome.}
	\vspace{-2mm}
	\label{fig:results}
\end{figure}

\noindent\textbf{Supervision ablation.}
Figure~\ref{fig:results}(a, d) tests whether learning to answer is enough to guide a search. Compared with Answer-only SFT, Trajectory SFT improves mean AQ from 58.4\% to 61.5\% and ESPA from 44.1\% to 51.7\%. Joint State Supervision raises both metrics to 73.8\%, outperforming Trajectory SFT by 12.3 and 22.1 percentage points, respectively. This demonstrates the benefit of supervising process-state updates beyond tool-use trajectories. Removing revision demonstrations lowers ESPA to 54.0\%, highlighting the value of revising explanations when intermediate stages remain unresolved. The supplement will report \textit{L1}--\textit{L4} scores for Answer-only SFT and Joint State Supervision and a comparison at matched observation budgets.

\noindent\textbf{Inference strategy ablation.}
Panels (b, e) evaluate evidence-use strategies under a fixed policy. Using either temporal path alone reduces both scores, indicating their complementary roles in identifying change evidence. Selecting views before updating memory lowers ESPA to 59.0\%, while fixing dates before target localization yields 64.5\%. Replacing process memory with a chronological log reduces ESPA to 54.8\%, and removing it entirely further lowers ESPA to 36.8\%. These results highlight the importance of updating process memory with new evidence to guide subsequent observations. Additional analyses of key-stage evidence recall, cross-date crop consistency, and irrelevant-view rates are provided in the supplement.

\noindent\textbf{Failure rates and error composition.}
Figure~\ref{fig:results}(c) shows that \textit{L4} has the highest failure rate (38.8\%), compared with 24.0\% for \textit{L3} and 17.7\% and 14.8\% for \textit{L2-T} and \textit{L2-S}. Although \textit{L2} and \textit{L3} can involve more tool calls, \textit{L4} asks the agent to recover intermediate stages and explain their order. Even with 77.6\% overall AQ, reconstructing long-term change remains the hardest task for the full model.

Of the 2,095 test answers, 470 did not meet the AQ criteria. We assign each to one category based on the main reason recorded by the scorer (Fig.~\ref{fig:results}(f)). The answers were incomplete in 218 cases (46.4\%), unresolved in 106 (22.6\%), and lacked a factual answer in 53 (11.3\%). The remaining 93 were clear errors, including incorrect change comparisons and misread turning points or cycles. Unresolved cases include unclear imagery, weak source evidence, or claims that could not be confirmed during extraction. Recovering all key stages and supporting them with verifiable evidence remains the main challenge.

\begin{figure}[t]
	\centering
	\includegraphics[width=0.99\columnwidth]{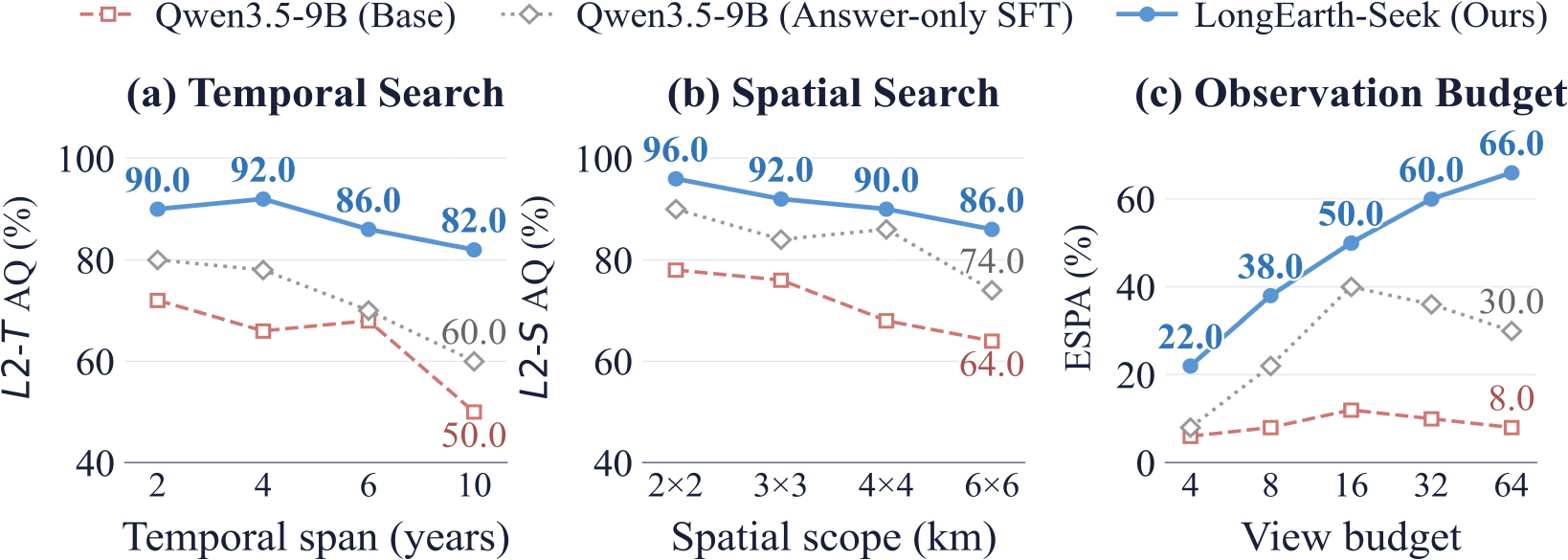}
	\vspace{-2mm}
	\caption{\textbf{Robustness and evidence use in process-guided search.} Each panel uses a fixed 50-question subset. (a) Temporal Search varies the accessible time span on \textit{L2-T}. (b) Spatial Search varies the geographic window on \textit{L2-S}. Both report AQ. (c) Observation Budget reports \textit{L4} ESPA at different view limits.}
	\label{fig:search_robustness}
	\vspace{-2mm}
\end{figure}

\noindent\textbf{Search robustness and evidence use.}
Figure~\ref{fig:search_robustness} varies the time span, search area, or view budget on fixed 50-question subsets. Even the smallest time and space windows retain enough evidence to answer each question. From 2 to 10 years, \textit{L2-T} AQ changes from 90.0\% to 82.0\% for LongEarth-Seek, from 72.0\% to 50.0\% for the base model, and from 80.0\% to 60.0\% for Answer-only SFT. In the widest window, \textit{L2-S} AQ is 86.0\% for LongEarth-Seek versus 74.0\% for Answer-only SFT. On \textit{L4}, raising the cap from 16 to 64 views lifts its ESPA from 50\% to 66\%, while Answer-only SFT falls from 40\% to 30\%.

\subsection{Qualitative Example}
\label{sec:qualitative}
\begin{figure}[t]
	\centering
	\vspace{-5mm}
	\includegraphics[width=0.98\columnwidth]{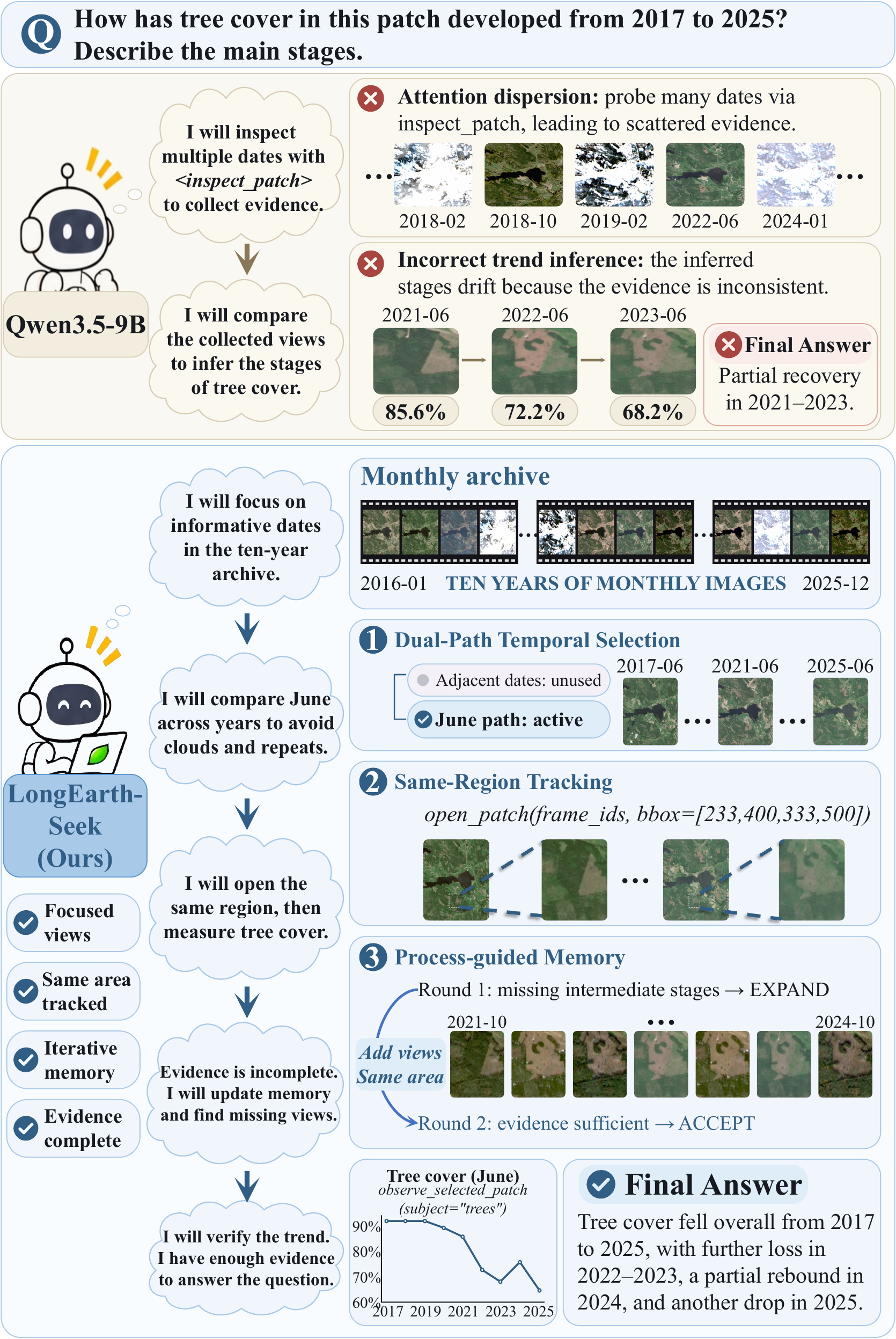}
	\caption{\textbf{How process-guided search traces tree-cover change.} Qwen3.5-9B mistakes decline for recovery. LongEarth-Seek recovers intermediate stages with memory and aligned views.}
	\label{fig:tree_cover_case}
	\vspace{-4mm}
\end{figure}

Figure~\ref{fig:tree_cover_case} compares two searches for the history of tree cover in the same patch from 2017 to 2025. Qwen3.5-9B inspects several dates, including cloudy views, and gathers scattered evidence. Its estimates fall from 85.6\% in June 2021 to 68.2\% in June 2023, yet its answer reports a partial recovery over that interval. The case shows how scattered observations can lead to an interpretation that drifts from the evidence. 
LongEarth-Seek also compares June images across years and opens the same patch at selected dates. When the first pass leaves intermediate stages unclear, the process memory directs the agent to inspect more views of that patch. The resulting measurements show further loss in 2022--2023, a partial rebound in 2024, and another drop in 2025. Each added view helps resolve a specific gap in the history of the same target.

\section{Conclusion}
\label{sec:conclusion}

We introduced LES-Bench to evaluate long-term surface change understanding and LongEarth-Seek to actively retrieve missing spatiotemporal evidence. By identifying unresolved change relations, maintaining spatial consistency across dates, and updating process memory with new observations, LongEarth-Seek integrates evidence acquisition with process reasoning. Experiments demonstrate improved evidence-grounded reasoning and generalization to broader geospatial tasks. Future work will integrate complementary evidence from optical, radar, and other geospatial observations with spatiotemporal reasoning to discover and reconstruct more complex geographic processes involving interacting changes and long-term dependencies.
{\small\bibliographystyle{ieeenat_fullname}\bibliography{main}}

@article{sentinel2,
  title = {{Sentinel-2: ESA's Optical High-Resolution Mission for GMES Operational Services}},
  author = {M. Drusch and U. Del Bello and S. Carlier and O. Colin and V. Fernandez and others},
  year = {2012},
  journal = {Remote Sensing of Environment},
  volume = {120},
  pages = {25--36},
  doi = {10.1016/j.rse.2011.11.026},
}

@article{dynamicworld,
  title = {{Dynamic World, Near real-time global 10 m land use land cover mapping}},
  author = {Christopher F. Brown and Steven P. Brumby and Brookie Guzder-Williams and Tanya Birch and Samantha Brooks Hyde and others},
  year = {2022},
  journal = {Scientific Data},
  volume = {9},
  pages = {251},
  doi = {10.1038/s41597-022-01307-4},
}

@article{jrcwater,
  title = {{High-resolution mapping of global surface water and its long-term changes}},
  author = {Jean-François Pekel and Andrew Cottam and Noel Gorelick and Alan S. Belward},
  year = {2016},
  journal = {Nature},
  volume = {540},
  pages = {418--422},
  doi = {10.1038/nature20584},
}

@article{hansen,
  title = {{High-Resolution Global Maps of 21st-Century Forest Cover Change}},
  author = {M. C. Hansen and P. V. Potapov and R. Moore and M. Hancher and S. A. Turubanova and others},
  year = {2013},
  journal = {Science},
  volume = {342},
  pages = {850--853},
  doi = {10.1126/science.1244693},
}

@article{levircc,
  title = {{Remote Sensing Image Change Captioning With Dual-Branch Transformers: A New Method and a Large Scale Dataset}},
  author = {Chenyang Liu and Rui Zhao and Hao Chen and Zhengxia Zou and Zhenwei Shi},
  year = {2022},
  journal = {IEEE Transactions on Geoscience and Remote Sensing},
  volume = {60},
  pages = {1--20},
  doi = {10.1109/TGRS.2022.3218921},
}

@article{changeagent,
  title = {{Change-Agent: Toward Interactive Comprehensive Remote Sensing Change Interpretation and Analysis}},
  author = {Chenyang Liu and Keyan Chen and Haotian Zhang and Zipeng Qi and Zhengxia Zou and Zhenwei Shi},
  year = {2024},
  journal = {IEEE Transactions on Geoscience and Remote Sensing},
  volume = {62},
  pages = {1--16},
  doi = {10.1109/TGRS.2024.3425815},
}

@article{remoteclip,
  title = {{RemoteCLIP: A Vision Language Foundation Model for Remote Sensing}},
  author = {Fan Liu and Delong Chen and Zhangqingyun Guan and Xiaocong Zhou and Jiale Zhu and Qiaolin Ye and Liyong Fu and Jun Zhou},
  year = {2024},
  journal = {IEEE Transactions on Geoscience and Remote Sensing},
  volume = {62},
  pages = {1--16},
  doi = {10.1109/TGRS.2024.3390838},
}

@article{internvl3,
  title = {{InternVL3: Exploring Advanced Training and Test-Time Recipes for Open-Source Multimodal Models}},
  author = {Zhu, Jinguo and Wang, Weiyun and Chen, Zhe and Liu, Zhaoyang and Ye, Shenglong and others},
  year = {2025},
  journal = {arXiv preprint arXiv:2504.10479},
}

@inproceedings{lora,
  title = {{LoRA: Low-Rank Adaptation of Large Language Models}},
  author = {Hu, Edward J. and Shen, Yelong and Wallis, Phillip and Allen-Zhu, Zeyuan and Li, Yuanzhi and Wang, Shean and Wang, Lu and Chen, Weizhu},
  year = {2022},
  booktitle = {ICLR},
}

@inproceedings{zoomearth,
  title = {{ZoomEarth: Active Perception for Ultra-High-Resolution Geospatial Vision-Language Tasks}},
  author = {Ruixun Liu and Bowen Fu and Jiayi Song and Kaiyu Li and Wanchen Li and Lanxuan Xue and Hui Qiao and Weizhan Zhang and Deyu Meng and Xiangyong Cao},
  year = {2026},
  booktitle = {CVPR},
  pages = {34877--34888},
}

@inproceedings{earthdial,
  title = {{EarthDial: Turning Multi-sensory Earth Observations to Interactive Dialogues}},
  author = {Sagar Soni and Akshay Dudhane and Hiyam Debary and Mustansar Fiaz and Muhammad Akhtar Munir and others},
  year = {2025},
  booktitle = {CVPR},
  pages = {14303--14313},
}

@inproceedings{videotree,
  title = {{VideoTree: Adaptive Tree-based Video Representation for LLM Reasoning on Long Videos}},
  author = {Ziyang Wang and Shoubin Yu and Elias Stengel-Eskin and Jaehong Yoon and Feng Cheng and Gedas Bertasius and Mohit Bansal},
  year = {2025},
  booktitle = {CVPR},
  pages = {3272--3283},
}

@inproceedings{act2see,
  title = {{Act2See: Emergent Active Visual Perception for Video Reasoning}},
  author = {Martin Q. Ma and Yuxiao Qu and Aditya Agrawal and Willis Guo and Paul Pu Liang and Ruslan Salakhutdinov and Louis-Philippe Morency},
  year = {2026},
  booktitle = {CVPR},
  pages = {5455--5464},
}

@inproceedings{longvideor1,
    author    = {Qiu, Jihao and Xie, Lingxi and Huo, Xinyue and Tian, Qi and Ye, Qixiang},
  title = {{LongVideo-R1: Smart Navigation for Low-cost Long Video Understanding}},
    booktitle = {CVPR},
    month     = {June},
    year      = {2026},
    pages     = {40505-40515},
  url = {https://openaccess.thecvf.com/content/CVPR2026/html/Qiu_LongVideo-R1_Smart_Navigation_for_Low-cost_Long_Video_Understanding_CVPR_2026_paper.html}
}

@inproceedings{geochat,
  title = {{GeoChat: Grounded Large Vision-Language Model for Remote Sensing}},
  author = {Kartik Kuckreja and Muhammad Sohail Danish and Muzammal Naseer and Abhijit Das and Salman Khan and Fahad Shahbaz Khan},
  year = {2024},
  booktitle = {CVPR},
  pages = {27831--27840},
}

@inproceedings{skysenseo,
  title = {{SkySense-O: Towards Open-World Remote Sensing Interpretation with Vision-Centric Visual-Language Modeling}},
  author = {Qi Zhu and Jiangwei Lao and Deyi Ji and Junwei Luo and Kang Wu and others},
  year = {2025},
  booktitle = {CVPR},
  pages = {14733--14744},
}

@inproceedings{skysense,
  title = {{SkySense: A Multi-Modal Remote Sensing Foundation Model Towards Universal Interpretation for Earth Observation Imagery}},
  author = {Xin Guo and Jiangwei Lao and Bo Dang and Yingying Zhang and Lei Yu and others},
  year = {2024},
  booktitle = {CVPR},
  pages = {27672--27683},
}

@inproceedings{videomme,
  title = {{Video-MME: The First-Ever Comprehensive Evaluation Benchmark of Multi-modal LLMs in Video Analysis}},
  author = {Chaoyou Fu and Yuhan Dai and Yongdong Luo and Lei Li and Shuhuai Ren and others},
  year = {2025},
  booktitle = {CVPR},
  pages = {24108--24118},
}

@inproceedings{longvideobench,
  title = {{LongVideoBench: A Benchmark for Long-context Interleaved Video-Language Understanding}},
  author = {Haoning Wu and Dongxu Li and Bei Chen and Junnan Li},
  year = {2024},
  booktitle = {NeurIPS},
}

@inproceedings{qlora,
  title = {{QLoRA: Efficient Finetuning of Quantized LLMs}},
  author = {Tim Dettmers and Artidoro Pagnoni and Ari Holtzman and Luke Zettlemoyer},
  year = {2023},
  booktitle = {NeurIPS},
}

@misc{qwen35,
  title = {{Qwen3.5: Towards Native Multimodal Agents}},
  author = {{Qwen Team}},
  year = {2026},
  howpublished = {Official model release and model card},
  url = {https://huggingface.co/Qwen/Qwen3.5-9B},
}

@misc{gemini31,
  title = {{Gemini 3.1 Pro Model Card}},
  author = {{Google DeepMind}},
  year = {2026},
  howpublished = {Official model card},
  url = {https://deepmind.google/models/model-cards/gemini-3-1-pro/},
}

@inproceedings{teochat,
  author = {Irvin, Jeremy Andrew and Liu, Emily Ruoyu and Chen, Joyce Chuyi and Dormoy, Ines and Kim, Jinyoung and others},
  title = {{TEOChat}: A Large Vision-Language Assistant for Temporal Earth Observation Data},
  booktitle = {ICLR},
  year = {2025},
  url = {https://openreview.net/forum?id=pZz0nOroGv}
}

@inproceedings{dynamicvl,
  author = {Xuan, Weihao and Wang, Junjue and Qi, Heli and Chen, Zihang and Zheng, Zhuo and Zhong, Yanfei and Xia, Junshi and Yokoya, Naoto},
  title = {{DynamicVL}: Benchmarking Multimodal Large Language Models for Dynamic City Understanding},
  booktitle = {NeurIPS},
  volume = {38},
  year = {2025},
  doi = {10.52202/085713-1114},
  url = {https://proceedings.neurips.cc/paper_files/paper/2025/hash/2fb80f455dc6056422c8af87dca50489-Abstract-Datasets_and_Benchmarks_Track.html}
}

@inproceedings{geochrono,
  author = {Li, Yujie and Pan, Jiancheng and Wei, Zhiwei and Wang, Jiuniu and Peng, Mugen and Xu, Wenjia},
  title = {{GeoChrono}: Benchmarking and Rethinking Long-Term Temporal Understanding in Remote Sensing},
  booktitle = {ACM MM},
  year = {2026},
  url = {https://arxiv.org/abs/2607.15768}
}

@article{rsagent,
  author = {Xu, Wenjia and Yu, Zijian and Mu, Boyang and Wang, Jiuniu and Wei, Zhiwei and Peng, Mugen},
  title = {{RS-Agent}: Automating Remote Sensing Tasks through Intelligent Agent},
  journal = {Science China Information Sciences},
  volume = {69},
  number = {8},
  pages = {180302},
  year = {2026},
  doi = {10.1007/s11432-026-5026-5}
}

@inproceedings{earthagent,
  author = {Feng, Peilin and Lv, Zhutao and Ye, Junyan and Wang, Xiaolei and Huo, Xinjie and others},
  title = {{Earth-Agent}: Unlocking the Full Landscape of Earth Observation with Agents},
  booktitle = {ICLR},
  year = {2026},
  url = {https://openreview.net/forum?id=dkIXAbWuxO}
}

@inproceedings{geommag,
  author = {Xiao, Aoran and Cheng, Shihao and Xu, Yonghao and Ren, Yexian and Chen, Hongruixuan and Yokoya, Naoto},
  title = {{GeoMMBench} and {GeoMMAgent}: Toward Expert-Level Multimodal Intelligence in Geoscience and Remote Sensing},
  booktitle = {CVPR},
  pages = {34843--34853},
  year = {2026},
  url = {https://openaccess.thecvf.com/content/CVPR2026/html/Xiao_GeoMMBench_and_GeoMMAgent_Toward_Expert-Level_Multimodal_Intelligence_in_Geoscience_and_CVPR_2026_paper.html}
}

@inproceedings{openearthagent,
  author = {Shabbir, Akashah and Sheikh, Muhammad Umer and Munir, Muhammad Akhtar and Debary, Hiyam and Fiaz, Mustansar and others},
  title = {{OpenEarthAgent}: A Unified Framework for Tool-Augmented Geospatial Agents},
  booktitle = {ECCV},
  pages = {172--191},
  year = {2026},
  doi = {10.1007/978-3-032-37574-2_10}
}

@inproceedings{hirsagent,
  author = {Mu, Boyang and Wei, Zhiwei and Peng, Mugen and Xu, Wenjia},
  title = {{HiRS-Agent}: A Hierarchical Multi-Agent System for Reliable Long-Horizon Remote Sensing Task Solving},
  booktitle = {ACM MM},
  year = {2026},
  pages = {7796--7805},
  doi = {10.1145/3767308.3835311},
  url = {https://doi.org/10.1145/3767308.3835311}
}

@inproceedings{univearth,
  author = {Kao, Chia Hsiang and Zhao, Wenting and Lam, Cheryl and Umap, Aarush and Revankar, Shreelekha and Speas, Samuel and Bhagat, Snehal and Datta, Rajeev and Phoo, Cheng Perng and Mall, Utkarsh and Vondrick, Carl and Bala, Kavita and Hariharan, Bharath},
  title = {Towards {LLM} Agents for Earth Observation},
  booktitle = {ACL Findings},
  pages = {2597--2611},
  year = {2026},
  doi = {10.18653/v1/2026.findings-acl.124},
  url = {https://aclanthology.org/2026.findings-acl.124/}
}

@article{vlrsbench,
  author = {Luo, Zhiming and Wang, Di and Wang, Hebaixu and Guo, Haonan and Zhang, Jing and Du, Bo},
  title = {{VLRS-Bench}: A Vision-Language Reasoning Benchmark for Remote Sensing},
  journal = {arXiv preprint arXiv:2602.07045v2},
  year = {2026},
  url = {https://arxiv.org/abs/2602.07045v2}
}

@inproceedings{activevideoperception,
  title = {{Active Video Perception: Iterative Evidence Seeking for Agentic Long Video Understanding}},
  author = {Ziyang Wang and Honglu Zhou and Shijie Wang and Junnan Li and Caiming Xiong and Silvio Savarese and Mohit Bansal and Michael S. Ryoo and Juan Carlos Niebles},
  year = {2026},
  booktitle = {CVPR Findings},
  pages = {9088--9099},
}

@inproceedings{videoarm,
  title = {{VideoARM}: Agentic Reasoning over Hierarchical Memory for Long-Form Video Understanding},
  author = {Yufei Yin and Qianke Meng and Minghao Chen and Jiajun Ding and Zhenwei Shao and Zhou Yu},
  booktitle = {CVPR},
  year = {2026},
  pages = {24042--24051}
}

@misc{gpt4o,
  author = {{OpenAI}},
  title = {{GPT-4o System Card}},
  year = {2024},
  url = {https://cdn.openai.com/gpt-4o-system-card.pdf}
}

@inproceedings{vitssits,
    author    = {Tarasiou, Michail and Chavez, Erik and Zafeiriou, Stefanos},
  title = {{ViTs for SITS: Vision Transformers for Satellite Image Time Series}},
    booktitle = {CVPR},
    month     = {June},
    year      = {2023},
    pages     = {10418-10428},
  url = {https://openaccess.thecvf.com/content/CVPR2023/html/Tarasiou_ViTs_for_SITS_Vision_Transformers_for_Satellite_Image_Time_Series_CVPR_2023_paper.html}
}

@inproceedings{segearthr2,
    author    = {Xin, Zepeng and Li, Kaiyu and Chen, Luodi and Li, Wanchen and Yuchen, Xiao and Qiao, Hui and Zhang, Weizhan and Meng, Deyu and Cao, Xiangyong},
  title = {{SegEarth-R2: Towards Comprehensive Language-guided Segmentation for Remote Sensing Images}},
    booktitle = {CVPR},
    month     = {June},
    year      = {2026},
    pages     = {13199-13210},
  url = {https://openaccess.thecvf.com/content/CVPR2026/html/Xin_SegEarth-R2_Towards_Comprehensive_Language-guided_Segmentation_for_Remote_Sensing_Images_CVPR_2026_paper.html}
}

@inproceedings{olmoearth,
    author    = {Herzog, Henry and Bastani, Favyen and Zhang, Yawen and Tseng, Gabriel and Redmon, Joseph and others},
  title = {{OlmoEarth: Stable Latent Image Modeling for Multimodal Earth Observation}},
    booktitle = {CVPR},
    month     = {June},
    year      = {2026},
    pages     = {34806-34817},
  url = {https://openaccess.thecvf.com/content/CVPR2026/html/Herzog_OlmoEarth_Stable_Latent_Image_Modeling_for_Multimodal_Earth_Observation_CVPR_2026_paper.html}
}

@inproceedings{robsense,
    author    = {Do, Minh Kha and Han, Kang and Lai, Phu and Phan, Khoa T. and Xiang, Wei},
  title = {{RobSense: A Robust Multi-modal Foundation Model for Remote Sensing with Static, Temporal, and Incomplete Data Adaptability}},
    booktitle = {CVPR},
    month     = {June},
    year      = {2025},
    pages     = {7427-7436},
  url = {https://openaccess.thecvf.com/content/CVPR2025/html/Kha_RobSense_A_Robust_Multi-modal_Foundation_Model_for_Remote_Sensing_with_CVPR_2025_paper.html}
}

@inproceedings{ramen,
    author    = {Houdr\'e, Nicolas and Marcos, Diego and de Turckheim, Hugo Riffaud and Ienco, Dino and Wendling, Laurent and Kurtz, Camille and Lobry, Sylvain},
  title = {{RAMEN: Resolution-Adjustable Multimodal Encoder for Earth Observation}},
    booktitle = {CVPR},
    month     = {June},
    year      = {2026},
    pages     = {27838-27848},
  url = {https://openaccess.thecvf.com/content/CVPR2026/html/Houdre_RAMEN_Resolution-Adjustable_Multimodal_Encoder_for_Earth_Observation_CVPR_2026_paper.html}
}

@inproceedings{terramind,
    author    = {Jakubik, Johannes and Yang, Felix and Blumenstiel, Benedikt and Scheurer, Erik and Sedona, Rocco and others},
  title = {{TerraMind: Large-Scale Generative Multimodality for Earth Observation}},
    booktitle = {ICCV},
    month     = {October},
    year      = {2025},
    pages     = {7383-7394},
  url = {https://openaccess.thecvf.com/content/ICCV2025/html/Jakubik_TerraMind_Large-Scale_Generative_Multimodality_for_Earth_Observation_ICCV_2025_paper.html}
}

@inproceedings{spectralchange,
    author    = {Zang, Qi and Zhao, Dong and Wang, Shuang and Quan, Dou and Zhong, Zhun},
  title = {{Feature Spectrum Learning for Remote Sensing Change Detection}},
    booktitle = {CVPR},
    month     = {June},
    year      = {2025},
    pages     = {12647-12657},
  url = {https://openaccess.thecvf.com/content/CVPR2025/html/Zang_Feature_Spectrum_Learning_for_Remote_Sensing_Change_Detection_CVPR_2025_paper.html}
}

@inproceedings{terrascope,
    author    = {Shu, Yan and Ren, Bin and Xiong, Zhitong and Zhu, Xiao Xiang and Demir, Beg\"um and Sebe, Nicu and Rota, Paolo},
  title = {{TerraScope: Pixel-Grounded Visual Reasoning for Earth Observation}},
    booktitle = {CVPR},
    month     = {June},
    year      = {2026},
    pages     = {16712-16722},
  url = {https://openaccess.thecvf.com/content/CVPR2026/html/Shu_TerraScope_Pixel-Grounded_Visual_Reasoning_for_Earth_Observation_CVPR_2026_paper.html}
}

@inproceedings{visprog,
    author    = {Gupta, Tanmay and Kembhavi, Aniruddha},
  title = {{Visual Programming: Compositional Visual Reasoning Without Training}},
    booktitle = {CVPR},
    month     = {June},
    year      = {2023},
    pages     = {14953-14962},
  url = {https://openaccess.thecvf.com/content/CVPR2023/html/Gupta_Visual_Programming_Compositional_Visual_Reasoning_Without_Training_CVPR_2023_paper.html}
}

@inproceedings{svagent,
    author    = {Yang, Zhongyu and Yang, Zuhao and Zhan, Shuo and Yue, Tan and Pang, Wei and Yuan, Yingfang},
  title = {{SVAgent: Storyline-guided Long Video Understanding via Cross-Modal Multi-Agent Collaboration}},
    booktitle = {CVPR},
    month     = {June},
    year      = {2026},
    pages     = {24062-24072},
  url = {https://openaccess.thecvf.com/content/CVPR2026/html/Yang_SVAgent_Storyline-guided_Long_Video_Understanding_via_Cross-Modal_Multi-Agent_Collaboration_CVPR_2026_paper.html}
}

@inproceedings{drvideo,
    author    = {Ma, Ziyu and Gou, Chenhui and Shi, Hengcan and Sun, Bin and Li, Shutao and Rezatofighi, Hamid and Cai, Jianfei},
  title = {{DrVideo: Document Retrieval Based Long Video Understanding}},
    booktitle = {CVPR},
    month     = {June},
    year      = {2025},
    pages     = {18936-18946},
  url = {https://openaccess.thecvf.com/content/CVPR2025/html/Ma_DrVideo_Document_Retrieval_Based_Long_Video_Understanding_CVPR_2025_paper.html}
}

@inproceedings{vipergpt,
    author    = {Sur{\'\i}s, D{\'\i}dac and Menon, Sachit and Vondrick, Carl},
  title = {{ViperGPT: Visual Inference via Python Execution for Reasoning}},
    booktitle = {ICCV},
    month     = {October},
    year      = {2023},
    pages     = {11888-11898},
  url = {https://openaccess.thecvf.com/content/ICCV2023/html/Suris_ViperGPT_Visual_Inference_via_Python_Execution_for_Reasoning_ICCV_2023_paper.html}
}

@inproceedings{moviechat,
    author    = {Song, Enxin and Chai, Wenhao and Wang, Guanhong and Zhang, Yucheng and Zhou, Haoyang and others},
  title = {{MovieChat: From Dense Token to Sparse Memory for Long Video Understanding}},
    booktitle = {CVPR},
    month     = {June},
    year      = {2024},
    pages     = {18221-18232},
  url = {https://openaccess.thecvf.com/content/CVPR2024/html/Song_MovieChat_From_Dense_Token_to_Sparse_Memory_for_Long_Video_CVPR_2024_paper.html}
}

@inproceedings{timechat,
    author    = {Ren, Shuhuai and Yao, Linli and Li, Shicheng and Sun, Xu and Hou, Lu},
  title = {{TimeChat: A Time-sensitive Multimodal Large Language Model for Long Video Understanding}},
    booktitle = {CVPR},
    month     = {June},
    year      = {2024},
    pages     = {14313-14323},
  url = {https://openaccess.thecvf.com/content/CVPR2024/html/Ren_TimeChat_A_Time-sensitive_Multimodal_Large_Language_Model_for_Long_Video_CVPR_2024_paper.html}
}

@inproceedings{salova,
    author    = {Kim, Junho and Kim, Hyunjun and Lee, Hosu and Ro, Yong Man},
  title = {{SALOVA: Segment-Augmented Long Video Assistant for Targeted Retrieval and Routing in Long-Form Video Analysis}},
    booktitle = {CVPR},
    month     = {June},
    year      = {2025},
    pages     = {3352-3362},
  url = {https://openaccess.thecvf.com/content/CVPR2025/html/Kim_SALOVA_Segment-Augmented_Long_Video_Assistant_for_Targeted_Retrieval_and_Routing_CVPR_2025_paper.html}
}

@inproceedings{rewind,
    author    = {Diko, Anxhelo and Wang, Tinghuai and Swaileh, Wassim and Sun, Shiyan and Patras, Ioannis},
  title = {{ReWind: Understanding Long Videos with Instructed Learnable Memory}},
    booktitle = {CVPR},
    month     = {June},
    year      = {2025},
    pages     = {13734-13743},
  url = {https://openaccess.thecvf.com/content/CVPR2025/html/Diko_ReWind_Understanding_Long_Videos_with_Instructed_Learnable_Memory_CVPR_2025_paper.html}
}

@inproceedings{mlvu,
    author    = {Zhou, Junjie and Shu, Yan and Zhao, Bo and Wu, Boya and Liang, Zhengyang and others},
  title = {{MLVU: Benchmarking Multi-task Long Video Understanding}},
    booktitle = {CVPR},
    month     = {June},
    year      = {2025},
    pages     = {13691-13701},
  url = {https://openaccess.thecvf.com/content/CVPR2025/html/Zhou_MLVU_Benchmarking_Multi-task_Long_Video_Understanding_CVPR_2025_paper.html}
}

@inproceedings{lvbench,
    author    = {Wang, Weihan and He, Zehai and Hong, Wenyi and Cheng, Yean and Zhang, Xiaohan and Qi, Ji and Ding, Ming and Gu, Xiaotao and Huang, Shiyu and Xu, Bin and Dong, Yuxiao and Tang, Jie},
  title = {{LVBench: An Extreme Long Video Understanding Benchmark}},
    booktitle = {ICCV},
    month     = {October},
    year      = {2025},
    pages     = {22958-22967},
  url = {https://openaccess.thecvf.com/content/ICCV2025/html/Wang_LVBench_An_Extreme_Long_Video_Understanding_Benchmark_ICCV_2025_paper.html}
}

@inproceedings{longvale,
    author    = {Geng, Tiantian and Zhang, Jinrui and Wang, Qingni and Wang, Teng and Duan, Jinming and Zheng, Feng},
  title = {{LongVALE: Vision-Audio-Language-Event Benchmark Towards Time-Aware Omni-Modal Perception of Long Videos}},
    booktitle = {CVPR},
    month     = {June},
    year      = {2025},
    pages     = {18959-18969},
  url = {https://openaccess.thecvf.com/content/CVPR2025/html/Geng_LongVALE_Vision-Audio-Language-Event_Benchmark_Towards_Time-Aware_Omni-Modal_Perception_of_Long_Videos_CVPR_2025_paper.html}
}
\end{document}